\documentclass[11pt,a4paper]{article}
\usepackage[hyperref]{acl2020}
\usepackage{times}
\usepackage{latexsym}
\usepackage{amsmath}
\usepackage{amssymb}
\usepackage{booktabs}
\usepackage{ifthen}
\usepackage{xspace}
\usepackage{multirow}
\usepackage{graphicx}
\usepackage{color,colortbl}
\usepackage{tabulary}
\usepackage{url}
\usepackage{soul}
\usepackage{titlesec}

\aclfinalcopy 
\def\aclpaperid{2902} 

\title{FEQA: A Question Answering Evaluation Framework for Faithfulness Assessment in Abstractive Summarization}

\author{Esin Durmus\thanks{\hspace{1ex}Most of the work is done while the authors were at Amazon Web Services AI.}\\ 
  Cornell University \\ 
  \texttt{ed459@cornell.edu} \\\And
  He He \\
  New York University \\
  \texttt{hehe@cs.nyu.edu} \\ \And 
  Mona Diab   \\ 
  The George Washington University
 \\
  \texttt{mtdiab@gwu.edu} \\
  }

\date{}

\newcommand\refsec[1]{Section~\ref{sec:#1}}

\newcommand\reffig[1]{Figure~\ref{fig:#1}}

\newcommand\reftab[1]{Table~\ref{tab:#1}}
\newcommand\refapp[1]{Appendix~\ref{sec:#1}}

\ifthenelse{\isundefined{\definition}}{\newtheorem{definition}{Definition}}{}
\ifthenelse{\isundefined{\assumption}}{\newtheorem{assumption}{Assumption}}{}
\ifthenelse{\isundefined{\hypothesis}}{\newtheorem{hypothesis}{Hypothesis}}{}
\ifthenelse{\isundefined{\proposition}}{\newtheorem{proposition}{Proposition}}{}
\ifthenelse{\isundefined{\theorem}}{\newtheorem{theorem}{Theorem}}{}
\ifthenelse{\isundefined{\lemma}}{\newtheorem{lemma}{Lemma}}{}
\ifthenelse{\isundefined{\corollary}}{\newtheorem{corollary}{Corollary}}{}
\ifthenelse{\isundefined{\alg}}{\newtheorem{alg}{Algorithm}}{}
\ifthenelse{\isundefined{\example}}{\newtheorem{example}{Example}}{}

\newcommand\ie{i.e.\xspace}
\newcommand\eg{e.g.\xspace}
\newcommand\nl[1]{\textit{``#1''}}

\newcommand\cnn{CNN/DM\xspace}
\newcommand\xs{XSum\xspace}

\newcommand\pgc{\textsc{pgc}\xspace}
\newcommand\fast{\textsc{FastRL}\xspace}
\newcommand\bottom{\textsc{BottomUp}\xspace}
\newcommand\tconv{\textsc{Tconv}\xspace}
\newcommand\bert{\textsc{BertSum}\xspace}

\newcommand\lead{\textsc{lead}\xspace}

\newcolumntype{L}[1]{>{\raggedright\let\newline\\\arraybackslash\hspace{0pt}}m{#1}}
\newcolumntype{C}[1]{>{\centering\let\newline\\\arraybackslash\hspace{0pt}}m{#1}}
\newcolumntype{R}[1]{>{\raggedleft\let\newline\\\arraybackslash\hspace{0pt}}m{#1}}

\begin{document}
\maketitle
\begin{abstract}
Neural abstractive summarization models are prone to generate content inconsistent with the source document, \ie \emph{unfaithful}.
Existing automatic metrics do not capture such mistakes effectively.
We tackle the problem of evaluating faithfulness of a generated summary given its source document.
We first collected human annotations of faithfulness for outputs from numerous models on two datasets.
We find that current models exhibit a trade-off between \emph{abstractiveness} and \emph{faithfulness}:
outputs with less word overlap with the source document are more likely to be unfaithful.
Next, we propose an automatic question answering (QA) based metric for faithfulness, FEQA,\footnote{\textbf{F}aithfulness \textbf{E}valuation with \textbf{Q}uestion \textbf{A}nswering.} which leverages recent advances in reading comprehension. 
Given question-answer pairs generated from the summary, a QA model extracts answers from the document;
non-matched answers indicate unfaithful information in the summary.
Among metrics based on word overlap, embedding similarity, and learned language understanding models,
our QA-based metric has significantly higher correlation with human faithfulness scores,
especially on highly abstractive summaries.

\end{abstract}
\section{Introduction}
\label{sec:intro}
 
Abstractive summarization models must aggregate salient content from the source document(s) and remain \emph{faithful}, \ie being factually consistent with information in the source documents.
Neural abstractive models are effective at identifying salient content and producing fluent summaries \cite{See2017GetTT,Chen2018FastAS,gehrmann-etal-2018-bottom}.
However, the generated summary may not always contain faithful information, which is vital for real-world applications.
\begin{table}[]
\centering
\begin{tabular}{>{\arraybackslash}m{7cm}}
\hline
\textbf{Source.} The world's oldest person has died a few weeks after 
celebrating her 117th birthday. \textcolor{red}{Born on March 5, 1898}, the 
great-grandmother had lived through two world wars, the 
invention of the television and the first successful powered 
aeroplane flight by the wright brothers...\\
\hline
\textbf{Output sentence.} The world 's oldest person has \textcolor{red}{died on March 5, 1898}.\\
\hline
\end{tabular}
    \caption{An example of \textit{unfaithful} output (highlighted in \textcolor{red}{red}); generated by \citet{gehrmann-etal-2018-bottom}.}
\label{tab:unfaithful_example_intro}
\end{table}
Table \ref{tab:unfaithful_example_intro} shows an example of unfaithful generation. 
Recent studies have shown that around $30\%$ of generated summaries contain unfaithful information \cite{cao2018faithful,falke2019ranking,kryscinski2019neural},
especially when the sentence combines content from multiple source sentences \cite{lebanoff2019analyzing}.

In this paper, we address the problem of evaluating \emph{faithfulness} of generated summaries given their source documents. Our key insight is that current models are limited by a trade-off between abstractiveness and faithfulness (\refsec{tradeoff}). On a wide range of systems and two datasets with varying levels of abstractiveness (\cnn and \xs),
we show that the number of unfaithful sentences (annotated by humans) increases as the summary becomes more abstractive (\ie less overlap with the source document). Next, we investigate a diverse set of existing automatic evaluation metrics
such as ROUGE, BERTScore~\cite{zhang2019bertscore}, and learned entailment models.
We find that their correlations with human scores of faithfulness drop significantly on highly abstractive summaries,
where deeper text understanding beyond surface similarity is needed.

Recently, question answering (QA) based automatic metrics have been proposed for evaluating content selection in summarization \cite{eyal2019question,scialom2019answers,chen2018semantic}.
Specifically, cloze-style QA is used to evaluate whether important information in the source is recovered from the summary.
Inspired by prior work,
we use \emph{automatically generated} QA pairs to represent information in the summary and validate it against the source.
Concretely, we generate a set of ``groundtruth'' QA pairs from the summary,
using a learned model that converts a declarative sentence and an answer span to a question (\refsec{qa-metric}).
Then, off-the-shelf reading comprehension models are evaluated on this set by extracting answer spans from the \emph{source documents}.
 High accuracy means that the summary and the source document tend to produce the same answers,
thus they are factually consistent with respect to the questions.
Compared to prior approaches using cloze tests, our question generation approach enables evaluation with a broader range of QA models and answer types (\eg extractive and generative), thus maximally taking advantage of progress in QA.

Among automatic metrics based on $n$-gram overlap, word embeddings, and language understanding models (relation extraction and entailment), FEQA has significantly higher correlation with human scores of faithfulness
and is the only metric that correlates with human scores on highly abstractive summaries from \xs.

\section{The Abstractiveness-Faithfulness Tradeoff}
\label{sec:tradeoff}
While extractive summarizers are largely faithful (since they copy sentences from the source document), current abstractive models struggle to produce faithful summaries without copying.
Similar to \citet{lebanoff2019analyzing}, we observe that factual errors occur more frequently as models generate more abstractive summary sentences, 
\ie less overlap with the source document.
In this section, we analyze generated summaries along two dimensions: abstractiveness and faithfulness.
Specifically, we aim to answer the following questions:
(1) How to quantify abstractiveness of a summary?
(2) Is abstractiveness encouraged more by the data or the model?
(3) How does being abstractive affect faithfulness?

\subsection{Characterizing Abstractiveness of a Summary}
\label{sec:abs-metrics}
Abstractive summarization involves rephrasing important content into brief statements,
ranging from minor editing of a source sentence to condensing multiple sentences in new words. 
Given a source document and a summary, we want to measure the level of abstractiveness of the summary.

Prior work measures abstractiveness by overlapped text spans between the summary and the document \cite{grusky2018newsroom,zhang2018abstractiveness},
or indirectly by the effectiveness of extractive baselines such as \lead-3 \cite{Nallapati2016SummaRuNNerAR}.
While metrics such as extractive fragment coverage and density \cite{grusky2018newsroom} provide a continuous measure of the level of abstractiveness, we define a more fine-grained categorization of abstractiveness by analyzing  how each sentence in the summary is formed. 

A more abstractive summary sentence aggregates content over a larger chunk of source text;
consequently it must copy fewer words to maintain brevity.
Therefore, we define the following abstractiveness types based on the amount of copying,
\eg copying a source sentence, one or more partial fragments from the source sentence, and individual words.
\begin{enumerate}
    \item \textbf{Sentence extraction:} the summary sentence is exactly the same as one of the source sentences.
 
\item \textbf{Span extraction:} the summary sentence is a substring of one of the source sentences,
\eg \nl{the plane was coming back from the NCAA final} is a span extracted from \nl{\ul{the plane was coming back from the NCAA final}, according to spokesman John Twork}.

\item \textbf{Word extraction:} the summary sentence is formed by a subset of the tokens in a source sentence,
\eg\nl{Capybara Joejoe has almost 60,000 followers} is a result of deleting words in \nl{\ul{Capybara Joejoe} who lives in Las Vegas \ul{has almost 60,000 followers} on Instagram}.

\item \textbf{Perfect fusion$_{k}$:} the summary sentence is constructed by piecing together the substrings from $k$ ($k>1$) source sentences in their original order, \eg\nl{Capybara Joejoe has almost 60,000 followers} is a perfect fusion of the sentences \nl{\ul{Capybara Joejoe} lives in Las vegas.} and  \nl{He \ul{has almost 60,000 followers} on Instagram.}

\end{enumerate}

To quantify the amount of abstractiveness of a set of summaries, we label each sentence with the first qualified type in the order above if it fits to one of these categories.

We then
define the score of each type as the percentage of sentences labeled by that category.
The types are ordered by increasing levels of abstractiveness.
For example, a summary with higher fusion scores and lower extraction scores is considered more abstractive.
In addition, we compute the percentage of \textbf{novel $n$-grams} that do not appear in the source document as another metric for abstractiveness. 

\subsection{Is abstractiveness from the model or the data?}
\label{sec:model-data}
Equipped with the metrics for abstractiveness above,
we want to further understand how abstractive the generated summaries are,
and whether the amount of abstractiveness is a result of the training data or the model.
Therefore, we compute abstractiveness scores for both the reference summaries and summaries generated from a diverse set of models on two datasets.

\paragraph{Datasets.} We use the CNN/DailyMail \cite{Hermann:2015:TMR:2969239.2969428,nallapati-etal-2016-abstractive} (\cnn) and the \xs \cite{xsum-emnlp} datasets,
which are both used for single-document news summarization tasks.
\cnn consists of articles from the CNN and Daily Mail websites, where the summaries comprise highlights in bullet points.
\xs consists of BBC articles, where the summaries comprise a single-sentence summary that is written as the opening introductory sentence for the article.
\xs was released in particular to promote research on highly abstractive summarization systems. \refapp{dataset-stats} provides statistics on \cnn and \xs datasets: they contain around 288k and 204k training examples, respectively; \cnn includes longer documents and summaries on average.  
\paragraph{Models.} Most neural abstractive summarization models are based on sequence-to-sequence models.
They differ in how summarization-specific operations such as copying/extraction are instantiated.
We consider 5 prominent models and 
summarize their characteristics in \reftab{model-comparison}.\footnote{We use state-of-the-art models proposed for each dataset at the time of writing.}
Details of each model can be found in \refapp{model-details}.
\textbf{\pgc} \cite{See2017GetTT} uses the copy mechanism during decoding to allow extraction.
\textbf{\fast} \cite{Chen2018FastAS} and \textbf{\bottom} \cite{gehrmann-etal-2018-bottom} decouple extraction and abstractive generation by learning to select sentences and words respectively in the first step;
this model has been shown to generate more abstractive summaries compared to \pgc.
\textbf{\tconv} \cite{xsum-emnlp} is initially designed for \xs, thus it does not include any explicit copying/extraction components and focuses on long text representation using convolutional neural networks.
\textbf{\bert} \cite{liu2019text} consists of a BERT-based encoder and a 6-layer Transformer decoder. 
It incorporates extraction implicitly by first fine-tuning the encoder on the extractive summarization task.\footnote{We use the \textsc{BertSumExtAbs} variation.}
\begin{center}
\begin{table}[t]
\centering
    \footnotesize{
\begin{tabular}{llll}
\toprule 
Systems & Extractor & Encoder & Decoder \\
\midrule 
\pgc & $-$ & LSTM & LSTM+copy \\
\fast & sentences & LSTM & LSTM+copy \\
\bottom & words & LSTM & LSTM+copy \\
\tconv & $-$ & CNN+topic & CNN \\
\bert &  $-$ & BERT-based  & Transformer\\
\bottomrule
\end{tabular}
}
\caption{Comparison of summarization systems in terms of model architecture.
}
\label{tab:model-comparison}
\end{table}
\end{center}
\begin{table*}
    \centering
    \footnotesize{
\begin{tabular}{|l|l|ccc|cc|ccc|} 
\hline
\multirow{2}{*}{Dataset} & \multirow{2}{*}{Model} & \multicolumn{3}{c|}{Extraction} & \multicolumn{2}{c|}{Perfect fusion} & \multicolumn{3}{c|}{Novel $n$-grams} \\
& & Sentence & Span & Word & $k=2$ & $k\geq2$ & $n=1$ & $n=2$ & $n=3$ \\
\hline
\multirow{4}{*}{\cnn} &
   Ref & $1.39$ & $2.14$ & $9.27$ & $12.92$ & $14.87$ & $12.40$ & $51.03$ & $71.22$\\
& \pgc & $35.45$ & $34.18$  & $15.45$  &  $10.90$ & $1.61$ & $0.62$ & $3.33$ & $7.42$ \\
& \fast & $8.94$  &  $40.06$  & $39.64$ & $4.22$ & $0.84$  & $0.82$ & $10.89$ & $20.74$ \\
& \bottom & $7.65$ & $17.98$ & $36.75$ & $21.86$ & $6.77$ & $0.86$ & $11.44$  & $22.40$  \\
& \bert &  $-$  & $13.73$ & $53.40$  & $16.18$ & $4.39$  & $5.23$ & $14.55$ & $23.09$ \\
\hline
\hline
\multirow{3}{*}{\xs} &
  Ref & $-$ & $-$ & $-$ & $0.87$  & $0.77$ & $39.20$ & $84.98$  & $96.05$ \\
& \pgc & $-$ & $-$ & $-$ & $0.41$ & $3.47$ & $30.08$ & $74.27$ &  $91.27$ \\
& \tconv & $-$  & $-$ & $-$ & $0.35$ & $2.31$  & $34.07$  & $80.62$  & $95.12$ \\
& \bert & $-$ & $-$  & $-$ & $0.33$ & $3.15$ & $28.93$ & $75.85$  &  $91.41$ \\
\hline
\end{tabular}
}
\caption{Abstractiveness measures of the models on CNN/DM and XSum datasets. The numbers for Extraction and Perfect fusion indicate \% of sentences generated with these strategies. Numbers for novel $n$-grams indicate \% of $n$-grams that are present in the output sentence but is not present in the source.}

\label{tab:abstractiveness_models}
\end{table*}

\textbf{Results.} Our goal is to understand the level of abstractiveness of summaries generated by different models,
and the influence on abstractiveness from the training data.
Therefore, we analyzed summaries generated by the above models on CNN/DM and XSum. We computed the metrics described in \refsec{abs-metrics} for both the generated summaries and the reference summaries on the test sets.
The results are shown in \reftab{abstractiveness_models}.

First, \textit{CNN/DM is more extractive than XSum}.
Extraction scores of the reference summaries in \cnn shows that almost half of the sentences are formed by deleting words in one of the source sentences.
This shows that sentence \textit{compression} \cite{knight2002summarization} is the main technique used for this dataset.
In contrast, none of the summary sentences in XSum are formed by copying from a single source sentence.
They are generated mostly by paraphrasing the input content, indicated by the large fraction of novel $n$-grams. 

Second, \textit{training data has a larger influence on the abstractiveness of model outputs}.
Similar to \citet{zhang2018abstractiveness}, we find that models trained on \cnn are near-extractive.  
However, the same models trained on \xs are significantly more abstractive.
In fact, none of the models produced any sentence that copies words/phrases from a single source sentence,
which is consistent with characteristics of the reference summaries in XSum.
The content is more often rephrased in novel words/phrases.
However, on both datasets, current models struggle to achieve the same level of abstractiveness as the reference summaries,
indicating that additional inductive bias is needed to condense multiple sentences by rephrasing. 

Third, \emph{different models have different ways of doing extraction}.
When trained on \cnn, \pgc generates the majority of sentences by copying complete source sentences,
whereas \fast, \bottom and \bert do simple compression by deletion more often.
In addition, \bottom does more fusion compared to \pgc, \fast and \bert.

\begin{figure*}[th]
    \includegraphics[width=\textwidth]{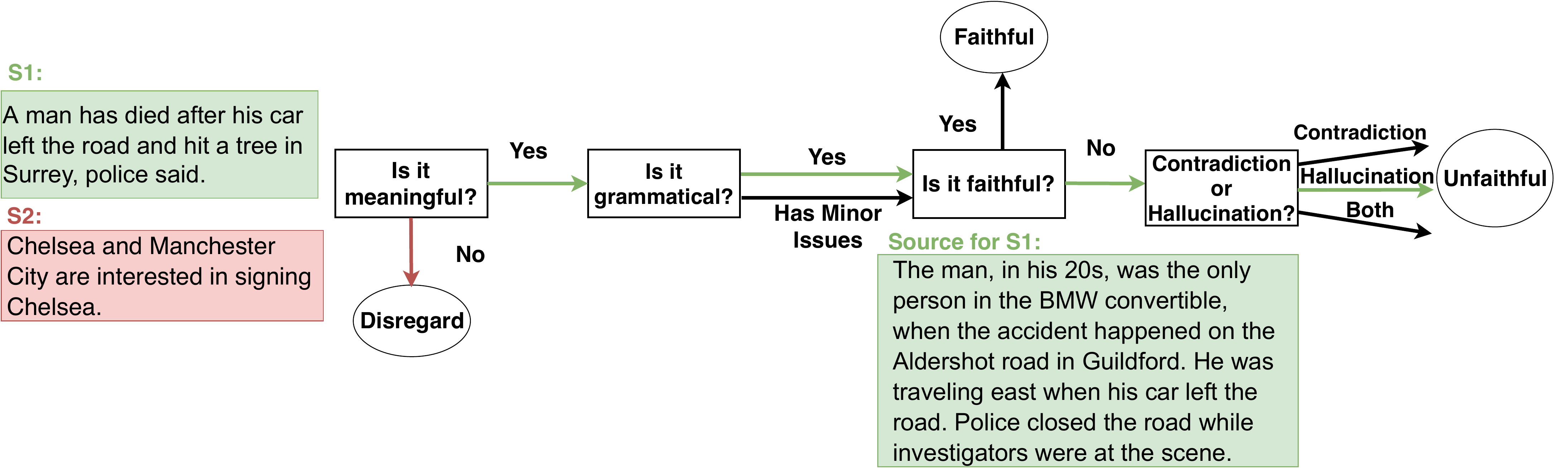}
    \caption{The decision diagram of our human annotation process. Decision nodes are rectangular and outcome nodes are circular. We show the annotation path of two summary sentences, S1 (green arrows) and S2 (red arrows).
    S2 is annotated as \textit{nonsensical} thus is not considered for faithfulness. S1 is annotated as unfaithful due to hallucinated content.
    }
\label{fig:dt-human-annotation}
\end{figure*}

\subsection{Annotating Summary Faithfulness\footnote{We make our data and code available for reproducibility at: \href{https://github.com/esdurmus/summary-faithfulness}{https://github.com/esdurmus/summary-faithfulness}.}} 
\label{sec:eval-faith}
To understand faithfulness of current systems and its relation to abstractiveness, we crowd-sourced human annotations on the output of each model-dataset pair described in Section \ref{sec:model-data}.
Since a near-extractive sentence is very likely to be grammatical and faithful,
we focus on more abstractive cases
by excluding output sentences that are either an exact copy or a substring of one of the source sentences. 

A key challenge to reliable human annotation is that the inter-annotator agreement on faithfulness is relatively low \cite{lebanoff2019analyzing}.
Our pilot study shows that workers often do not agree on incoherent sentences,
\eg whether \textit{``Chelsea beat Chelsea $5-3$ in the Premier League on Saturday.''} is faithful or not.
To standardize the annotation process, 
we design hierarchical questions to distinguish among
failed generation that render a sentence meaningless,
low-level grammatical errors that hardly affect semantic understanding,
and faithfulness errors that convey incorrect (yet meaningful) information.

\reffig{dt-human-annotation} shows the decision tree of our human annotation steps. 
We first evaluate the grammaticality of generated sentences (independent from the source document).
We show annotators a summary sentence and ask them to choose whether the given sentence is \textit{meaningful} or \textit{nonsensical} to determine if the given sentence is structurally and semantically sound.
If the annotator can make sense of the sentence, we then ask whether it is \textit{grammatical} or \textit{has minor grammaticality problems} which a person can easily correct.

Next, for sentences labeled as meaningful in the first step,
we ask workers whether they are faithful to the provided source document.
In case the worker labels a sentence as unfaithful, we conduct a simple error analysis by asking them to indicate if the sentence contains information that is absent from or conflicting with the source document,
which corresponds to  \emph{hallucination} and \emph{contradiction} errors, respectively.
More details about the annotation schema and guidelines are included in the \refapp{human-eval-details}.
Next, we describe our human evaluation results.

\begin{table*}
    \centering
    \footnotesize{
\begin{tabular}{|l|l|ccc|ccc|} 
\hline
\multirow{2}{*}{Dataset} & \multirow{2}{*}{Model} & \multicolumn{3}{c|}{Grammaticality} & \multicolumn{3}{c|}{ Faithfulness}  \\
&   &  Score & Agreement & Abstractiveness  & Score & Agreement & Abstractiveness\\
\hline
\multirow{3}{*}{\cnn} 
& \pgc   & $93.34$ & $94.04$ & $10.05$ & $70.05$ & $77.28$  & $13.35$\\
& \fast   & $83.06$ & $88.05$ & $44.46$ & $68.27$  & $77.45$ & $49.74$\\
& \bottom  & $85.83$ & $89.19$ & $29.62$ & $64.17$ & $76.04$  & $42.36$\\
& \bert   & $97.53$ & $97.65$ & $29.44$ & $95.03$ & $95.14$  & $39.16$\\
\hline

\hline
\multirow{2}{*}{\xs} 
& \pgc  & $65.85$ & $81.03$ & $91.10$ & $40.33$ & $71.63$  & $97.06$\\
& \tconv  & $70.85$ & $85.03$ & $94.94$ & $38.96$ & $69.90$ & $98.81$\\
& \bert  & $90.44$ & $91.80$ & $91.50$ & $60.54$ & $70.00$ & $97.60$\\

\hline
\end{tabular}
}
    \caption{Grammaticality and faithfulness results of human annotations. \textbf{Score} is computed by taking the percentage of  annotators that selected ``meaningful'' and ``faithful'' for grammaticality and faithfulness annotation tasks, respectively, and then averaging these values across all the examples for the given annotation task.
    \textbf{Agreement} is computed by taking the percentage of the workers that annotate the majority class for the given example. \textbf{Abstractiveness} is measured by the percentage of novel trigrams in a given sentence.}
\label{tab:human_eval_results}
\end{table*}

\subsubsection{Human Annotation Results}
\label{sec:human-eval-results}

For each dataset-model pair described in \refsec{model-data}, we randomly sampled $1000$ sentence-source pairs eliminating output sentences that are either an exact copy or substring of a source sentence. We collected grammaticality annotations for these sentences from $5$ annotators. 
We consider a sentence meaningful if at least $4$ out of $5$ annotators label it as meaningful in the first stage.
We sampled $200$ meaningful sentences randomly to collect annotations for  faithfulness. 
\reftab{human_eval_results} shows the results of the grammaticality and faithfulness human evaluations. 

\paragraph{Grammaticality.} Overall, outputs from all models are scored high on grammaticality with high inter-annotator agreement. However, on more \textit{abstractive} summaries (i.e. when trained on \xs), the grammaticality scores  drop significantly.
One exception is \bert, which maintains good performance on \xs and  achieves the highest grammaticality score on both datasets.\footnote{Majority of the sentences ($>70\%$) identified as ``meaningful'' are annotated as ``perfectly grammatical'' for each model-dataset pair.}
\paragraph{Faithfulness.} Near-extractive summaries generated from models trained on \cnn have significantly higher faithfulness scores than highly abstractive summaries from models trained on \xs. We find that \pgc and \tconv has faithfulness errors in more than half of the sentences they generate when trained on \xs. Although \bert generates fewer unfaithful sentences, it still suffers from performance drop on \xs. Interestingly, human agreement on faithfulness is also lower for abstractive summaries from \xs. This suggests that faithfulness errors are harder to catch for humans as well in more abstractive settings.
We further observe conflicting information is more common among models trained on \cnn while hallucination is more common among models trained on \xs. Table \ref{tab:example_unfaithful_main_text} shows examples of meaningful but unfaithful sentences. 
\begin{table*}
\centering
\footnotesize{
\begin{tabular}{>{\arraybackslash}m{5cm}>{\arraybackslash}m{4cm}>{\arraybackslash}m{2cm}>{\arraybackslash}m{2cm}}
\hline
\textbf{Source} & \textbf{Output Sentence} & \textbf{Domain} & \textbf{Category} \\ 
\hline
...However, \textbf{Winger Ross Wallace (knee) and right-back Steven Reid (calf)} could return for the Barclays premier league contest... & \textbf{Dean Marney and Steven Reid} could return for the Barclays Premier League match. & \cnn & IC\\ 
\hline
....Odom also played for the US in the 2004 Athens Olympics, winning the bronze medal. \textbf{His condition is unknown} but well-wishers tweeted their support following the news... & NBA basketball player Odom has been \textbf{found dead} in a helicopter crash in the US state of Nevada. & \xs & H\\ 
\hline
\end{tabular}
}
\caption{Examples of \textit{meaningful} but \textit{unfaithful} sentences. Category corresponds to the faithfulness \emph{error} type for the output sentence. \textbf{IC}: Incorrect Concatenation, \textbf{H}: Hallucination. More examples are provided in Table \ref{tab:example_unfaithful}. 
}
\label{tab:example_unfaithful_main_text}
\end{table*}
\begin{figure*}[th]
    \includegraphics[width=\textwidth]{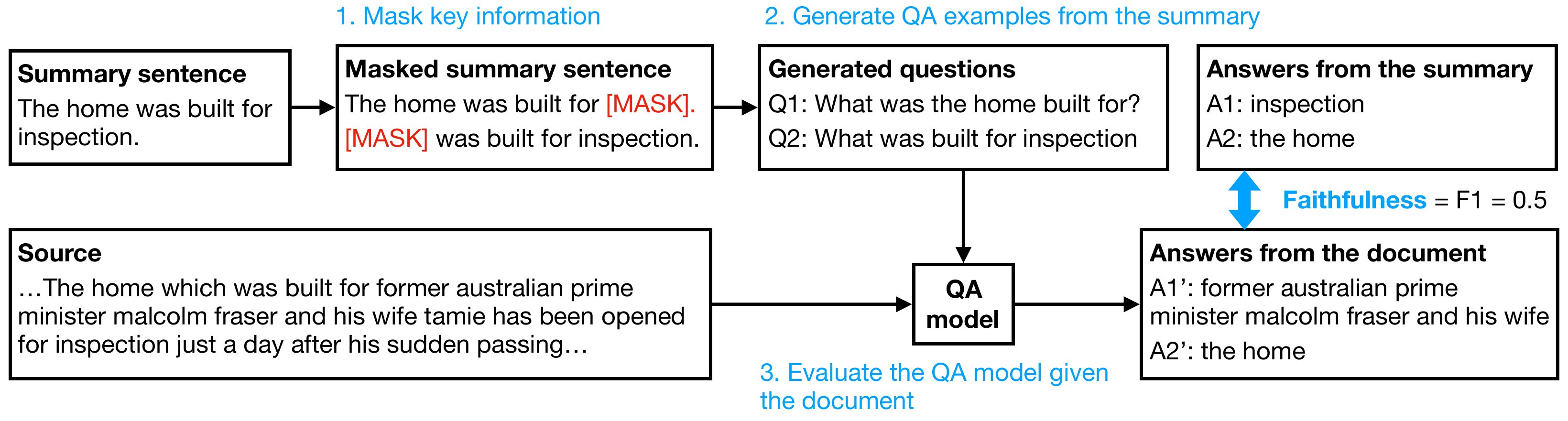}
    \caption{Overview of FEQA.
    Given a summary sentence and its corresponding source document, we first mask important text spans (\eg noun phrases, entities) in the summary.
    Then, we consider each span as the ``gold'' answer and
    generate its corresponding question using a learned model.
    Lastly, a QA model finds answers to these questions in the documents;
    its performance (\eg F1 score) against the ``gold'' answers from the summary is taken as the faithfulness score.
    }
\label{fig:qa-eval-overview}
\end{figure*}

\section{FEQA: Faithfulness Evaluation with Question Answering}
\label{sec:qa-metric}
Our analysis above shows that the number of unfaithful sentences increases significantly as  more abstractive summaries are generated.
Thus the key challenge to faithfulness evaluation is to verify highly abstractive sentences against the source document,
where surface similarity matching would fail. 
If we have a good semantic representation of the sentence abstracting away its surface form (\eg a list of facts about \emph{who} did \emph{what} to \emph{whom}),
we can simply compare the sentence representation to the document representation (\eg check whether the fact list from the summary is a subset of the list from the document).
Ideally, the representation should be domain-general and interpretable for easy error analysis.

Motivated by the fast progress in reading comprehension \cite{chen2018neural,gao-etal-2018-neural-approaches}
we propose to use QA pairs as a generic meaning representation of sentences for faithfulness evaluation.
Given a summary sentence, we produce a list of questions asking about key information in the sentence and their corresponding answers.
To verify this information against the source, 
we use a QA model to predict answers from the document.
The questions and the QA model thus extract comparable information from two pieces of text.
More matched answers from the document implies a more faithful summary
since the information addressing these questions are consistent between the summary and the source document. 
\reffig{qa-eval-overview} shows the workflow of FEQA.

\paragraph{Question generation.}
Prior work \cite{eyal2019question,scialom2019answers} uses cloze tests as questions by masking entities.
To go beyond cloze-style QA and leverage more recent extractive \cite{rajpurkar-etal-2016-squad} or even generative \cite{radford2019language} QA models,
we generate natural language questions from the summary sentence automatically.
Specifically, we mask important text spans in a sentence,
including noun phrases extracted by a constituency parser \cite{kitaev-klein-2018-constituency} and named entities extracted by the Stanford CoreNLP NER model \cite{Finkel:2005:INI:1219840.1219885,manning-etal-2014-stanford}.
We consider each span as the gold answer and generate its corresponding question by fine-tuning a pretrained BART language model \cite{Lewis2019BARTDS}.
To train the question generator, we adapt the QA2D dataset \citet{Demszky2018TransformingQA}.
The input is a declarative sentence with masked answers
and the output is a question.
A training example might look like:
\begin{table}[h]
\begin{tabular}{rl}
Input:&\texttt{\small Sally was born in <m> 1958 </m>} \\
Output:&\texttt{\small When was Sally born ?}
\end{tabular}
\end{table}

Since the transformation from declarative sentences to questions is almost rule-based without much paraphrasing,
we expect the model to generalize to various domains.

\paragraph{Answer verification.}
Given the QA pairs generated from a summary sentence, we run  off-the-shelf QA models to get answers to these questions from the source document.
We then measure the average F1 score against the ``gold'' answers from the summary, which is our faithfulness score for the given sentence.
This step does not have any constraint on the QA model.
We experiment with the pretrained BERT-\texttt{base} model \cite{Devlin2019BERTPO} fine-tuned on SQuAD-1.1 \cite{rajpurkar-etal-2016-squad} and SQuAD-2.0 \cite{rajpurkar-etal-2018-know}. Note that in the case of SQuAD-2.0, the model may be able to hypothesize that a question is unanswerable. This case is equivalent to getting an answer incorrect (i.e. unfaithful). 

\section{Experiments}
\label{sec:experiments}

We aim to understand to what extent the proposed QA-based metric and existing metrics capture faithfulness of a summary.
Given pairs of documents and summary sentences \emph{without reference summaries},
we measure correlations between human-annotated faithfulness scores (\refsec{eval-faith}) and scores computed using each metric described below. 

\subsection{Automated Metrics for Faithfulness}
\paragraph{Word overlap-based metrics.} \label{word-overlap-metric}
A straightforward metric for faithfulness is the word overlap between the summary sentence and the document.
We compute ROUGE (R), BLEU (B),\footnote{We report only BLUE-4 since it performed the best for \cnn and no variation of BLEU has significant correlation with faithfulness for \xs.} between the output sentence and each of the source sentences (i.e. taking the source sentence as the reference). We then take the average scores and maximum score  across all the source sentences. Since according to our analysis taking the average score consistently has higher correlation, we report only the correlation for the average.

\paragraph{Embedding-based metrics.}\label{embedding-metric}
Word embeddings extend word overlap-based metrics beyond exact match.
Recently, BERTScore \cite{Zhang2019BERTScoreET} was proposed to compute the similarity between two sentences using contextual word embeddings from BERT. 
It has higher correlation with human judgements on image captioning and machine translation than word overlap based metrics.
We compute BERTScore (BERTSc) between each source sentence and the summary sentence.\footnote{\href{https://github.com/Tiiiger/bert_score}{https://github.com/Tiiiger/bert\_score}.}
To get the final score, we experiment with both the average and the maximum scores computed from each source sentence and the summary sentence. We report results using the maximum score since it has better performance. 

\paragraph{Model-based metrics.}\label{model-based-metric}
In addition to QA, recent work has used relation extraction and textual entailment models for faithfulness evaluation \cite{falke2019ranking,goodrich2019assessing}.
For the relation extraction metric (RE), we compute the precision for the relation triplets extracted from the summary sentence and the source document using an off-the-shelf model \cite{angeli-etal-2015-leveraging} from Stanford Open IE.
For the textual entailment metric (ENT), we measure whether the summary sentence is entailed by the source using the pretrained ESIM model \cite{chen-etal-2017-enhanced} from AllenNLP \cite{gardner-etal-2018-allennlp}.

\subsection{Results}
\begin{table*}[]
\centering
    \footnotesize{
\begin{tabular}{>{\arraybackslash}m{4cm}>{\arraybackslash}m{4.5cm}>{\arraybackslash}m{2.5cm}>{\arraybackslash}m{1.5cm}>{\arraybackslash}m{1.5cm}}
Source Sentence & Output Sentence & Metric & Score  \\
\hline
Health Inspectorate Wales said Wrexham Maelor Hospital staff were under ``considerable pressure'' for long periods as ambulances waited outside.  & A hospital ward in Wrexham has been rated ``inadequate'' by inspectors after inspectors found patients at risk of harm.
& 
Entailment &  $72.83\%$
\\ 
\hline
The Black Poplar is one of the rarest native trees in the UK, with only 2,500 thought to be left. & Northern Ireland's first trees are among those recognised in the Welsh Architecture Trust's list of the year's best trees.  
& BertScore & $83.06\%$
 \\ 
\hline
\end{tabular}
}
\caption{Unfaithful examples missed by Entailment and BertScore. \textbf{Score}: Output score of the metrics; higher score indicates stronger entailment and similarity respectively.} 
\label{tab:metric-comp-ex}
\end{table*}
\paragraph{Metric Comparison.}
We first compute scores for each metric on document and output sentence pairs on both \cnn and \xs datasets ($748$ and $286$ pairs respectively).
We then compute Pearson and Spearman correlation coefficients between scores given by each metric and human-annotated scores.
\reftab{metric_correlations} includes correlation coefficients for the examples from \cnn and \xs, respectively. We observe that for both \cnn and \xs, the score of QA-based evaluation has a higher correlation with faithfulness than other metrics. Although word-overlap based metrics are correlated with the faithfulness in more extractive settings (i.e. for \cnn), these metrics have no correlation with faithfulness in more abstractive settings (i.e. for \xs). We further notice that all the metrics have significantly lower correlation with human scores for \xs, suggesting that evaluating faithfulness is more difficult in highly abstractive settings; deeper understanding of the source and the summary sentence is necessary here. 

Consistent with the findings of \citet{falke-etal-2019-ranking}, the entailment metric does not have a significant correlation with faithfulness in most cases. These models fail to distinguish entailed (faithful) and non-entailed (unfaithful) summary sentences when both overlap largely with the source document,
because models trained on current entailment datasets may rely on simple heuristics such as lexical overlap \cite{mccoy2019hans}. Similarly, BERTScore tends to give higher scores when there are overlapping concepts between the sentences even though the content is not the same. See 
\reftab{metric-comp-ex} for examples.

\begin{table}
\centering
    \footnotesize{
\begin{tabular}{|l|rr|rr|} 
\hline
 &  \multicolumn{2}{c|}{\cnn} & \multicolumn{2}{c|}{\xs}  \\
Metric & P & S  & P  & S \\
\hline
\multicolumn{5}{|c|}{Word overlap-based} \\
\hline
$\text{R-1}$ 
& $12.02
^{**}$ & $15.86^{**}$  
& $-2.57$ & $0.07$\\ 
$\text{R-2}$ 
& $13.25^{**}$ & $15.99^{**}$  & $-5.78$ & $-8.47$\\
$\text{R-L}$  
& $12.58^{**}$& $16.49^{**}$ & $-6.37$ & $-9.68$
\\ 
$\text{B-4}$  
& $12.09^{**}$ & $11.68^{**}$ 
& $-6.76^{  }$ & $-10.02^{  }$\\
\hline 
\multicolumn{5}{|c|}{Embedding-based} \\
\hline
$\text{BERTSc}$  & $11.07^{*}$ & $10.70^{*}$& $10.06$ & $10.69$\\
\hline 
\multicolumn{5}{|c|}{Model-based} \\
\hline
    $\text{RE}$  & $8.58^{*\hphantom{*}}$ & $5.52^{\hphantom{**}}$ & $1.62^{\hphantom{**}  }$ & $2.32^{\hphantom{**}}$ \\ 
    $\text{ENT}$  & $2.80^{\hphantom{**}}$ & $3.65^{\hphantom{**}}$& $-5.62^{\hphantom{**}}$& $-3.85^{\hphantom{**}}$\\ 
$\text{FEQA}$ & $\textbf{32.01}^{**}$ & $\textbf{28.23}^{**}$ & $\textbf{26.31}^{**}$ & $\textbf{21.34}^{**}$\\ 
\hline
\end{tabular}
}
\caption{Pearson (P) and Spearman (S) correlation between human-annotated \textit{faithfulness scores} and the metric scores. *,** indicates $p$-values $<0.05$,$<0.001$, respectively. FEQA has the highest correlation with human scores for both \cnn and \xs.}
\label{tab:metric_correlations}
\end{table}

\paragraph{Content selection and faithfulness.}
Current evaluation metrics for summarization produce a single measure of the overall quality of the summary.
Typically, the output summary is compared against the reference summary in terms of n-gram overlap.
These metrics mainly evaluate \emph{content selection}, \ie whether the content of the output is similar to the content of the reference.
In contrast, to evaluate faithfulness, we compare the output summary against the source document. One natural question that follows is whether high content matching sufficient for faithfulness.
We compute the correlation coefficients between human-annotated faithfulness scores
and ROUGE scores computed from the reference and the output sentence.
As shown in \reftab{content_selection_correlation}, 
while there is a weak correlation between ROUGE scores of content selection and faithfulness on \cnn,
the correlation is significantly lower than ROUGE scores of faithfulness (i.e. computed between the \textit{source} and the output sentence).
For \xs, there is no significant correlation between the content selection metrics and faithfulness.
We provide unfaithful examples with high content selection scores in \refapp{high_content_unfaithful}. 
This suggests that {content selection} and {faithfulness} should be measured separately as opposed to using a unified score. 

\begin{table}
\centering
    \footnotesize{
\begin{tabular}{|l|ll|rr|} 
\hline
 &  \multicolumn{2}{c|}{\cnn} & \multicolumn{2}{c|}{\xs}  \\
Metric & P & S  & P  & S \\
\hline

$\text{ROUGE-1}$  
& $15.31^{**}$ & $14.92^{**}$
& $5.44$ & $5.79$\\
$\text{ROUGE-2}$  
& $15.10^{**}$ & $16.39^{**}$ & $8.25$ & $6.79$\\
$\text{ROUGE-L}$ 
& $13.33^{**}$ & $13.35^{**}$ 
& $4.61$ & $3.97$\\
\hline
\end{tabular}
}
\caption{Pearson (P) and Spearman (S) correlation between human-annotated faithfulness scores and ROUGE scores of content selection (computed between the reference and the output sentence). High content selection scores (typical ROUGE score for summarization) do not necessarily imply faithfulness of the summary.}
\label{tab:content_selection_correlation}
\end{table}
\begin{table*}[]
\centering
    \footnotesize{
\begin{tabular}{>{\arraybackslash}m{4cm}>{\arraybackslash}m{3.5cm}>{\arraybackslash}m{2.5cm}>{\arraybackslash}m{1.5cm}>{\arraybackslash}m{1.5cm}}
Source & Output Sentence & Question & OA & SA  \\
\hline
...However, Winger Ross Wallace (knee) and right-back Steven Reid (calf) could return for the Barclays premier league contest... & \textbf{Dean Marney} and Steven Reid could return for the Barclays Premier League match. & Who and Steven Reid could return for the premier league match? & Dean Marney
& Ross Wallace\\ 
\hline
 ...Miss Bruck, 22, from maybe has not been seen since the early hours of October 26, 2014. She has not been seen for six months...  & Miss Bruck, 22, from maybe has not been seen for \textbf{six months.} & How long has Miss Bruck, 22 from not been seen for? & six months & six months\\ 
\hline
\end{tabular}
}
    \caption{Examples detection results from FEQA. \textbf{OA}:Output Answer, \textbf{SA}:Source Answer. The output sentence in the first example is \textit{unfaithful}, whereas the one for the second example is \textit{faithful}. Bold text indicates the span that was masked to generate the question.} 
\label{tab:example_qa_correct}
\end{table*}

\paragraph{Analysis and limitations of QA-based evaluation.} \reftab{example_qa_correct} shows examples for a faithful and an unfaithful output sentence and the corresponding QA pairs. Note that the QA system is able to capture common errors such as conflicting information in the output sentence. 
To measure the reliability of FEQA, we further perform a manual error analysis using $100$ randomly sampled QA pairs.  We observe that around $94\%$ of generated questions are mostly grammatical and correct given the mask. 
For $78\%$ of the questions, the QA system has the correct behaviour: it answers the question correctly if the sentence is faithful to the article, otherwise it produces ``unanswerable'' or an incorrect answer.  Majority of the errors of the QA system are because it either didn't detect unanswerable questions or produces ``unanswerable'' when there exists an answer ($14\%$). Moreover, when the article is long, QA system tends to make more mistakes. Especially for more abstractive settings, F1-score penalizes the correct answers when the answer from the article does not exactly match with the gold answer (i.e. ``Donald Trump'' vs. ``the President of the United States Donald Trump'') ($16\%$).

\section{Related Work}
\label{sec:related}
\paragraph{Problems in current neural generation models.}
Since the beginning of neural text generation, problems with repetition and generic responses have received lots of attention \cite{sordoni2015neural,li2016diversity,holtzman2019curious}.
Recently, more work has focused on semantic errors in model outputs, such as adequacy in machine translation \cite{tu2017neural}, faithfulness in summarization \cite{cao2018faithful}, and consistency in dialogue \cite{li2019dont}.
Our analysis on the abstractiveness-faithfulness tradeoff reveals additional limitation of current models,
and suggests that we need new inductive bias on how to summarize beyond copying.

\paragraph{QA as a proxy.}
Question answering is a broad format that subsumes many tasks \cite{gardner2019question}.
To the best of our knowledge, \citet{mani1999tipster} first use QA as an extrinsic evaluation for summarization: A good summary should answer key questions a reader might have about an article.
Later, QA is incorporated in human evaluation where one person writes questions and another person answers them based on the summary \cite{clarke2010discourse,liu2019text}.
The closest to our work are recent efforts in automating this protocol,
including rule-based approaches \cite{chen2018semantic}
and cloze-test QA \cite{eyal2019question,scialom2019answers}.
Our work is the first to apply automated question generation.
While we focus on faithfulness, our QA-based metric is applicable to semantic comparison between any two pieces of text.

\paragraph{Automated evaluation for NLG.}
Automated NLG evaluation is challenging as it often requires deep understanding of the text.
Although metrics based on word overlap with the reference text are commonly used,
it is widely known that they do not correlate well with human judgments \cite{novikova2017why,liu2016evaluate}.
Recently, more work has focused on model-based evaluation using discriminators \cite{lowe2017towards,zhang2019huse},
entailment models \cite{falke2019ranking}, information extraction \cite{wiseman2017challenges,goodrich2019assessing}, and question answering \cite{chen2018semantic,eyal2019question}.

\section{Conclusion}
We investigate the faithfulness problem in neural abstractive summarization
and propose a QA-based metric for evaluating summary faithfulness.
We show that current models suffer from an inherent trade-off between abstractiveness and faithfulness.
They are good at copying important source content, but
tend to concatenate unrelated spans and hallucinate details when generating more abstractive sentences.
A new inductive bias or additional supervision is needed for learning reliable models.
While our QA-based metric correlates better with human judgment and is useful for model development,
it is limited by the quality of the QA model.
The final evaluation should still rely on human annotation or human-in-the-loop methods \cite{chaganty2018evaluation}.

\bibliography{all,acl2020}
\bibliographystyle{acl_natbib}
\clearpage
\appendix
\section*{Acknowledgement}
We would like to thank Faisal Ladhak, the Lex and Comprehend groups at Amazon Web Services AI, and the anonymous reviewers for their feedback on this work.

\section{Summarization Datasets}
\label{sec:dataset-stats}
All of our experiments are run on the \cnn and \xs datasets.
We show basic statistics of the two datasets in \reftab{basic_stats_datasets}.

\begin{table}[ht]
    \centering
    \footnotesize{
 \begin{tabular}{|l | r | r|} 
\hline
 & CNN/DM & XSum\\ [0.5ex] 
 \hline
 \# Training Documents & 287,227 &   204,045 \\ 
 \# Validation Documents & 13,368  &   11,332 \\
 \# Test Documents & 11,490   & 11,334\\
 \hline
 Document: avg \# of tokens & 781.00  & 431.07\\
  Document: avg \# of sents. & 40.00  & 33.00\\
 Summary: avg \# tokens & 56.00 & 23.26 \\ 
 Summary: avg \# of sents. & 3.75 & 1.00 \\ 
 \hline
\end{tabular}
}
\caption{Statistics of CNN/DM and XSum datasets.}
\label{tab:basic_stats_datasets}
\end{table}

\section{Summarization Models}
\label{sec:model-details}
The characteristics of each model used in our experiments are detailed below.

\paragraph{Pointer Generator Model with Coverage (\pgc) \cite{See2017GetTT}} uses the copy mechanism \cite{NIPS2015_5866} to allow copying words from the source.
The adapted \textit{coverage} mechanism \cite{tu-etal-2016-modeling} is incorporated to alleviate repetition by keeping track of source words that have been summarized.
This copy mechanism is widely adopted by subsequent models.

\paragraph{Fast Abstractive Summarization with
Reinforce (\fast) \cite{Chen2018FastAS}} first uses an extractor agent to select salient sentences from the document,
then condenses the extracted sentences using the Pointer-Generator summarizer. 

\paragraph{Bottom-up Summarization Model (\bottom) \cite{gehrmann-etal-2018-bottom}}
first selects words from the source document that are likely to appear in the summary,
then generates using the Pointer-Generator model,
where the copying mechanism is constrained to the previously selected words.
It improves upon \pgc by explicitly learning the selector to avoid copying long text spans.


\paragraph{Topic-aware Convolutional Sequence-to-Sequence model ({\sc tconvs2s}) \cite{xsum-emnlp}} is a convolutional neural network-based model conditioned on the topics of the article.
It is shown to be effective in capturing long-range dependencies in the documents.

\paragraph{BERT-based model (\bert) \cite{liu2019text}} is a two-stage fine-tuning approach where the BERT-based encoder is first fine-tuned on the extractive summarization task and then on the abstractive sumarization task with the decoder (denoted as \textsc{BertSumExtAbs} in the original paper).

\section{Details of Human Annotations}
\label{sec:human-eval-details}

\subsection{Grammaticality Annotation Guidelines}
For grammaticality annotation, we present only the output sentence to the workers. We collect annotations from $5$ workers for both of the tasks. For this task, given the output sentence, we provide workers the following guidelines:

\begin{enumerate}
  \item First select whether the given sentence is ``Nonsensical'' or ``Makes sense''. 
  \item If the given text is not a complete sentence, mark it as ``Nonsensical''.
  \item If you can understand the meaning of the sentence, despite  grammaticality errors,  and you are able to makes sense of it, select ``Makes sense''.
  \item If you did not select ``Nonsensical'', evaluate whether the sentence is ``Grammatical'' or ``Has Minor Grammaticality Issues''. 
\end{enumerate}

\subsection{Faithfulness Annotation Guidelines}

We present workers both the source and the output sentence and provide the following guidelines:
\begin{enumerate}
  \item Read the sentence and the source fully. 
  \item If the information conveyed by the sentence is not expressed in the source, select ``unfaithful''.
  \item Avoid using general knowledge, and check if the sentence is consistent with the source. 
  \item If you select ``unfaithful'', for the second part, select whether the information expressed by the sentence is \textit{not contained} in the source or \textit{conflicting }with the source. 
\end{enumerate}

\begin{table*}
\centering
\begin{tabular}{>{\arraybackslash}m{7cm}>{\arraybackslash}m{5cm}>{\arraybackslash}m{2cm}}
\hline
\textbf{Source} & \textbf{Output Sentence} & \textbf{Category} \\ 

\hline

...Although her due date has not officially been confirmed, the duchess of Cambridge told wellwishers at a charity event last month: \textbf{I am due mid-April, to the end of April}... &  The duchess of Cambridge told wellwishers at a charity event last month: \textbf{``The duke’s intention is to be at the commemorations''.} &  IC\\  
\hline

...Carragher spoke to a local TV starton during his time in Girona. Carragher posted a picture on his Instagram account\textbf{ of the opening ceremony}... & Carragher posted a picture \textbf{on his son play in the famous youth tournament}. &   IC\\  

\hline

 A body was found by a member of the public on private land near Leighton, about 10 miles (16.09km) away from the centre of Shrewsbury, on Monday. Mr Bebbington's family has been informed, West Mercia Police confirmed.  & The death of a man whose body was found in a river in Cumbria \textbf{has been identified as murder.}  & H\\
 \hline

The incident happened near Dr. Gray's hospital shortly after 10:00. The man was taken to the hospital with what police said were serious but not life-threatening injuries. The a96 was closed in the area for several hours, but it has since reopened. & A man has been taken to hospital after \textbf{he was hit by a lorry in Dumfries.} & H \\  
\hline
\end{tabular}
\caption{Examples of \textit{meaningful} but \textit{unfaithful} sentences. Category corresponds to the category of unfaithfulness error for the output sentence. \textbf{IC}: Incorrect Concatenation, \textbf{H}: Hallucination.}
\label{tab:example_unfaithful}
\end{table*}
\begin{table*}[]
\centering
\begin{tabular}{>{\arraybackslash}m{7cm}>{\arraybackslash}m{6cm}}
\hline
\textbf{Reference} & \textbf{Output Sentence}\\ 
\hline
... University of Nebraska researcher has revealed why stress is bad for you.
Limited periods of stress are good, as they release cortisol... &   University of Nebraska researcher has revealed why stress is bad for you, stimulating your body to produce an important hormone called cortisol.\\ 
\hline
...Indian air force and Nepalese army medical team launch rescue mission to bring injured people to hospitals in Kathmandu. Forshani Tamang's family carried her for four hours to reach help after she was wounded when their home was destroyed...  & Indian air crew and Nepalese army medical team were killed in Nepal's Sindhupalchok quake. \\
\hline
\end{tabular}
\caption{Examples of unfaithful sentence with high content overlap (computed by ROUGE-L) with the reference.}
\label{tab:example_high_overlap}
\end{table*}

\section{Additional Analysis}
\subsection{Examples for \textit{nonsensical} sentences}
\begin{itemize}
  \item Sandals, £34, office.co.uk, luluguinness.com. (generated by \pgc for \cnn)
  \item He says easter triduum is a progression , although the word itself -- triduum. (generated by \fast for \cnn)
  \item Chelsea beat Chelsea $5-3$ in the Premier League on Saturday. (generated by \fast for \cnn)
  \item 12 years a slave actress Lupita Woodley and oily vegetables. (generated by \bottom for \cnn)
  \item A judge in Japan has ordered a judge to order a woman who has absconded from Japan to Japan. (generated by \pgc for \xs)
  \item Stoke City moved up to third in the Premier League with victory over Stoke City at Stoke. (generated by \tconv for \xs) 
  \item Johnny Depp's management group is suing his management group over his ``lavish lifestyle''. (generated by \bert for \xs)

\end{itemize}

\begin{table*}[]
\centering
\begin{tabular}{>{\arraybackslash}m{7cm}>{\arraybackslash}m{6cm}}
\hline
\textbf{Source} & \textbf{Reference}\\ 
\hline
Ms Wood blamed the Conservatives in particular for claiming the SNP posed a threat to the future of the UK. She claimed "progressive" parties like hers were offering a ``collaborative'' alternative to ``combative'' politics. ``This election presents an opportunity for harmonious co-existence between our nations,'' she said. Ms Wood's comments followed Conservative claims that Labour dependence on support from the SNP to form a government after the election on 7 May would threaten the break-up of the UK. Campaigning in south Wales on Monday, she said: ``The parties advocating progressive, inclusive non-partisan cooperation in this election are not those who claim to cherish the political union above all others, but the national parties of Wales and Scotland. Along with the Greens in England, our parties have provided people across these islands with a collaborative alternative to the traditional combative Westminster politics.''. Ms Wood added that she had received ``hundreds'' of supportive messages from people in England following the televised debates.
& \textbf{Plaid Cymru leader Leanne} Wood has accused rival parties of "dangerous and divisive rhetoric" in a "desperate" attempt to win votes.\\
\hline

\hline
\end{tabular}
\caption{Example where reference includes information that \textbf{is not} in the source. }
\label{tab:dataset_issues}
\end{table*}

\subsection{Examples for \textit{meaningful} but \textit{unfaithful} sentences}

\reftab{example_unfaithful} includes examples that are annotated as \textit{meaningful} but \textit{unfaithful}. First three examples are picked from the models trained on \cnn, and last three are from the models trained on \xs. We observe that majority of sentences with faithfulness errors for \cnn dataset are generated by incorrect concatenation (IC). The models fuse two sentences from the source and generate a new sentence that is not consistent with the context of the source. Within this category, however, the models make a wide-range of mistakes such as copying the wrong entity, date, and quote.

For \xs, the faithfulness mistakes are mostly hallucinations. Models tend to hallucinate information (e.g. entities, events, date) that is not present in the source. 

\subsection{Examples for sentences with high content overlap with reference that are \textit{unfaithful}}\label{sec:high_content_unfaithful}
Although current summarization models are evaluated with respect to the content overlap between the reference and the output, these metrics do not necessarily provide any guarantees for the faithfulness of the output. \reftab{example_high_overlap} includes examples with similar content overlap scores as the faithful examples but are unfaithful. We can see that although the output sentences include similar words and refer to  similar topics, they include hallucinations and inaccurate information.



\subsection{Limitations of the datasets}

Since \cnn and \xs datasets are automatically crawled, we find that there is noise in the data. For example, source documents can include phrases such as ``click here for the latest news''. We further observe that \textit{reference} can carry information that is not in the source document since some of these one sentence highlights are written using additional world knowledge. \reftab{dataset_issues} shows an example where the reference is unfaithful since it includes information that is not in the source (i.e. the fact that Ms. Wood's first name is Leanne and she is Plaid Cymru leader.).

\end{document}